\documentclass[colacl]{article}
\usepackage{colacl}                              
\usepackage{times}                               

\title{An XML-based document suite}

\author{Dietmar R\"osner \and Manuela Kunze\\
Otto-von-Guericke-Universit\"at Magdeburg\\Institut f\"ur Wissens-
und Sprachverarbeitung \\
P.O.box 4120, 39016 Magdeburg,
Germany\\(roesner,makunze)@iws.cs.uni-magdeburg.de\\}

\begin{document}
\maketitle
\begin{abstract} We report about the current state of
development of a document suite and its applications.  This
collection of tools for the flexible and robust processing of
documents in German is based on the use of XML as unifying
formalism for encoding input and output data as well as process
information. It is organized in modules with limited
responsibilities that can easily be combined into pipelines to
solve complex tasks. Strong emphasis is laid on a number of
techniques to deal with lexical and conceptual gaps that are
typical when starting a new application. \end{abstract}



\newtheorem{example}{Example }

\section*{Introduction}
We have designed and implemented the XDOC document suite as a
workbench for the flexible processing of electronically available
documents in German. We have decided to exploit XML
\cite{xml-standard} and its accompanying formalisms (e.g. XSLT
\cite{XSL}) and tools (e.g. xt \cite{clarkSite} ) as a unifying
framework. All modules in the XDOC system expect XML documents as
input and deliver their results in XML format.

XML -- and ist precursor SGML -- offers a formalism to annotate
pieces of (natural language) texts. To be more precise: If a text
is (as a simple first approximation) seen as a sequence of
characters (alphabetic and \hyphenation{white-space} whitespace
characters) then XML allows to associate arbitrary markup with
arbitrary subsequences of {\em contiguous} characters.  Many
linguistic units of interest are represented by strings of
contiguous characters (e.g. words, phrases, clauses etc.). To use
XML to encode information about such a substring of a text
interpreted as a meaningful linguistic unit and to associate this
information directly with the occurrence of the unit in the text
is a straightforward idea. The basic idea is further backed by
XMLs demand that XML elements have to be properly nested. This is
fully concordant with standard linguistic practice: complex
structures are made up from simpler structures covering substrings
of the full string in a nested way.

The end users of our applications are domain experts (e.g. medical
doctors, engineers, ...). They are interested in getting their
problems solved but they are typically neither interested nor
trained in computational linguistics. Therefore the barrier to
overcome before they can use a computational linguistics or text
technology system should be as low as possible.

This experience has consequences for the design of the document
suite. The work in the XDOC project is guided by the following
design principles that have been abstracted from a number of
experiments and applications with "realistic" documents (i.a.
emails, abstracts of scientific papers, technical documentation,
...):

\begin{itemize}
  \item The tools shall be usable for `realistic' documents.
  \newline
  One aspect of `realistic' documents is that they typically contain
domain-specific tokens that are not directly covered by classical
lexical categories (like noun, verb, ...). Those tokens are
nevertheless often essential for the user of the document (e.g. an
enzyme descriptor like EC 4.1.1.17 for a biochemist).
  \item The tools shall be as robust as possible.
\\In general it can not be expected that lexicon information is
available for all tokens in a document. This is not only the case
for most tokens from `nonlexical' types -- like telephone numbers,
enzyme names, material codes, ... --, even for lexical types there
will always be `lexical gaps'. This may either be caused by
\hyphenation{neo-logisms} neologisms or simply by starting to
process documents from a new application domain with a new
sublanguage. In the latter case lexical items will typically be
missing in the lexicon (`lexical gap') and phrasal structures may
not or not adequately be covered by the grammar.
  \item The tools shall be usable independently but shall allow for
flexible combination and interoperability.
  \item The tools shall not only be usable by developers but as well by
domain experts without linguistic training.
\end{itemize}

Here again XML and XSLT play a major role: XSL stylesheets can be
exploited to allow different presentations of internal data and
results for different target groups; for end users the internals
are in many cases not helpful, whereas developers will need them
for debugging.

\newpage
The tools in the XDOC document suite can be grouped according to
their function:

\begin{itemize}
  \item preprocessing
  \item structure detection
  \item POS tagging
  \item syntactic parsing
  \item semantic analysis
  \item tools for the specific application: e.g. information extraction
\end{itemize}

In all tools the results of processing is encoded with XML tags
delimiting the respective piece of text. The information conveyed
by the tag name is enriched with XML attributes and their resp.
values.

\section*{Preprocessing} Tools for preprocessing are used to
convert documents from a number of formats into the XML format
amenable for further processing. As a subtask this includes
treatment of special characters (e.g. for umlauts, apostrophes,
...).

\section*{Structure detection}

We accept raw ASCII texts without any markup as input. In such
cases structure detection tries to uncover linguistic units (e.g.
sentences, titles, ...) as candidates for further analysis. A
major subtask is to identify the role of interpunction characters.

If we have the structures in a text explicitly available this may
be exploited by subsequent linguistic processing. An example: For
a unit classified as title or subtitle you will accept an NP
whereas within a paragraph you will expect full sentences.

In realistic texts even the detection of possible sentence
boundaries needs some care. A period character may not only be
used as a full stop but may as well be part of an abbreviation
(e.g. `z.B.' -- engl.: `e.g.' -- or `Dr.'), be contained in a
number (3.14), be used in an email address or in domain specific
tokens. The resources employed are special lexica (e.g. for
abbreviations) and finite automata for the reliable detection of
token from specialized non-lexical categories (e.g. enzyme names,
material codes, ...).

These resources are used here primarily to identify those full
stop characters that function as sentence delimiters (tagged as
IP). In addition, the information about the function of strings
that include a period is tagged in the result (e.g. ABBR).

\begin{example} results of structure detection
\scriptsize
\begin{verbatim}
Anwesend<IP>:</IP>
<ABBR>Univ.-Prof.</ABBR>
<ABBR>Dr.</ABBR><ABBR>med.</ABBR>Dieter Krause<IP>,</IP>
Direktor des Institutes fuer Rechtsmedizin
\end{verbatim}
\end{example}
\normalsize


\section*{POS tagging}

To try to assign part-of-speech information to a token is not only
a preparatory step for parsing. The information gained about a
document by POS tagging and evaluating its results is valuable in
its own right. The ratio of token not classifiable by the POS
tagger to token classified may e.g. serve as an indication of the
degree of lexical coverage.

In principle a number of approaches is usable for POS tagging
(e.g. \cite{brill:92}). We decided to avoid approaches based on
(supervised) learning from tagged corpora, since the cost for
creating the necessary training data are likely to be prohibitive
for our users (especially in specialized sublanguages).

The approach chosen was to try to make best use of available
resources for German and to enhance them with additional
functionality. The tool chosen is not only used in POS tagging but
serves as a general morpho-syntactic component for German:
MORPHIX.

The resources employed in XDOC's POS tagger are:

- the lexicon and the inflectional analysis from the
morphosyntactic component MORPHIX

- a number of heuristics (e.g. for the classification of token not
covered  in the lexicon)

For German the morphology component MORPHIX
\cite{finkler.neumann:88} has been developed in a number of
projects and is available in different realisations. This
component has the advantage that the closed class lexical items of
German as well as all irregular verbs are covered. The coverage of
open class lexical items is dependent on the amount of lexical
coding. The paradigms for e.g. verb conjugation and noun
declination are fully covered but to be able to analyze and
generate word forms their roots need to be included in the MORPHIX
lexicon.

We exploit MORPHIX - in addition to its role in syntactic parsing
- for POS tagging as well. If a token in a German text can be
morphologically analysed with MORPHIX the resulting word class
categorisation is used as POS information.  Note that this
classification need not be unique. Since the tokens are analysed
in isolation multiple analyses are often the case. Some examples:
the token `der' may either be a determiner (with a number of
different combinations for the features case, number and gender)
or a relative pronoun, the token `liebe' may be either a verb or
an adjective (again with different feature combinations not
relevant for POS tagging).

In addition since we do not expect extensive lexicon coding at the
beginning of an XDOC application some tokens will not get a
MORPHIX analysis. We then employ two techniques: We first try to
make use of heuristics that are based on aspects of the tokens
that can easily be detected with simple string analysis (e.g.
upper-/lowercase, endings, ...) and/or exploitation of the token
position relative to sentence boundaries (detected in the
structure detection module). If a heuristic yields a
classification the resulting POS class is added together with the
name of the employed heuristic (marked as feature SRC, cf. example
3). If no heuristics are applicable we classify the token as
member of the class unknown (tagged with XXX).

To keep the POS tagger fast and simple the disambiguation between
multiple POS classes for a token and the derivation of a possible
POS class from context for an unknown token are postponed to
syntactic processing. This is in line with our general principle
to accept results with overgeneration when a module is applied in
isolation (here: POS tagging) and to rely on filtering ambiguous
results in a later stage of processing (here: exploiting the
syntactic context).

\begin{example} domain-specific tagging

\scriptsize
\begin{verbatim}

<PRODUCT Method="Sandguss" Material="CC333G">
    <N>Gussstueck</N>
    <NORM>
         <N>EN</N>
         <NR>1982</NR>
    </NORM>
    <IP>-</IP>
    <MAT-ID>CC333G</MAT-ID>
    <IP>-</IP>
    <METHODE>GS</METHODE>
    <IP>-</IP>
    <MODELLNR>XXXX</MODELLNR>
</PRODUCT>
\end{verbatim}
\end{example}
\normalsize

The example above is the result of tagging a domain-specific
identifier. The token is annotated as a {\em PRODUCT} with
description of the used method and material. It is a typical token
in the domain of casting technology.
\section*{Syntactic parsing}

For syntactic parsing we apply a chart parser based on context
free grammar rules augmented with feature structures.

Again robustness is achieved by allowing as input elements:
\begin{itemize}
  \item multiple POS classes,
  \item unknown classes of open world tokens and
  \item tokens with POS class, but without or only partial feature
information.
\end{itemize}

\begin{example} unknown token classified as noun with heuristics
\scriptsize
\begin{verbatim}
<NP TYPE="COMPLEX" RULE="NPC3" GEN="FEM"
        NUM="PL" CAS="_">
  <NP TYPE="FULL" RULE="NP1" CAS="_"
            NUM="PL" GEN="FEM">
       <N SRC="UNG">Blutanhaftungen</N>
  </NP>
  <PP CAS="DAT">
    <PRP CAS="DAT">an</PRP>
    <NP TYPE="FULL" RULE="NP2" CAS="DAT"
            NUM="SG" GEN="FEM">
      <DETD>der</DETD>
      <N SRC="UC1">Gekroesewurzel</N>
    </NP>
  </PP>
</NP>
\end{verbatim}
\end{example}
\normalsize

The latter case results from some heuristics in POS
tagging that allow to assume e.g. the class noun for a token but
do not suffice to detect its full paradigm from the token (note
that there are ca two dozen different morphosyntactic paradigms
for noun declination in German).

For a given input the parser attempts to find all complete
analyses that cover the input. If no such complete analysis is
achievable it is attempted to combine maximal partial results into
structures covering the whole input.

A successful analysis may be based on an assumption about the word
class of an initially unclassified token (tagged XXX). This is
indicated in the parsing result (feature AS) and can be exploited
for learning such classifications from contextual constraints.  In
a similar way the successful combination from known feature values
from closed class items (e.g. determiners, prepositions) with
underspecified features in agreement constraints allows the
determination of paradigm information from successfully processed
occurrences. In example 4 features of the unknown word
"Mundhoehle" could be derived from the features of the determiner
within the PP.

\begin{example} unknown token classified as adjective
and features derived through contextual constraints
\scriptsize
\begin{verbatim}
<NP TYPE="COMPLEX" RULE="NPC3" GEN="MAS" NUM="SG"
    CAS="NOM">
  <NP TYPE="FULL" RULE="NP3" CAS="NOM" NUM="SG"
        GEN="MAS">
    <DETI>kein</DETI>
    <XXX AS="ADJ">ungehoeriger</XXX>
    <N>Inhalt</N>
  </NP>
  <PP CAS="DAT">
    <PRP CAS="DAT">in</PRP>
    <NP TYPE="FULL" RULE="NP2" CAS="DAT" NUM="SG"
        GEN="FEM">
      <DETD>der</DETD>
      <N SRC="UC1">Mundhoehle</N>
    </NP>
  </PP>
</NP>"
\end{verbatim}
\end{example}
\normalsize The grammar used in syntactic parsing is organized in
a modular way that allows to add or remove groups of rules. This
is exploited when the sublanguage of a domain contains linguistic
structures that are unusual or even ungrammatical in standard
German.

\begin{example}Excerpt from syntactic analysis
\scriptsize
\begin{verbatim}
<PP CAS="AKK">
  <PRP CAS="AKK">durch</PRP>
    <NP TYPE="COMPLEX" RULE="NPC1" GEN="NTR" NUM="SG"
                                 CAS="AKK">
      <NP TYPE="FULL" RULE="NP1" CAS="AKK" NUM="SG"
                                 GEN="NTR">
        <N>Schaffen</N>
      </NP>
      <NP TYPE="FULL" RULE="NP2" CAS="GEN" NUM="SG"
                                  GEN="MAS">
         <DETD>des</DETD>
         <N>Zusammenhalts</N>
      </NP>
    </NP>
</PP>
\end{verbatim}
\end{example}

\newpage
\section*{Semantic analysis}

At the time of writing semantic analysis uses three methods:

\subsection*{Semantic tagging}

For semantic tagging we apply a semantic lexicon. This lexicon
contains the semantic interpretation of a token and a case frame
combined with the syntactic valence requirements. Similar to POS
tagging the tokens are annotated with their meaning and a
classification in semantic categories like e.g. concepts and
relations. Again it is possible, that the classification of a
token in isolation is not unique. Multiple classification can be
resolved through the following analysis of the case frame and
through its combination with the syntactic structure which
includes the token.

\subsection*{Analysis of case frames}

By the case frame analysis of a token we obtain details about the
type of recognized concepts (resolving multiple interpretations)
and possible relations to other concepts. The results are tagged
with XML tags. The following example describes the DTD for the
annotation of the results of case frame analysis.

\begin{example} DTD for the annotation by case frame analysis
\scriptsize
\begin{verbatim}
    <!ELEMENT CONCEPTS (CONCEPT)*>

    <!ELEMENT CONCEPT (WORD, DESC, SLOTS?)>
    <!ATTLIST CONCEPT TYPE CDATA #REQUIRED>

    <!ELEMENT WORD (#PCDATA)>
    <!ELEMENT DESC (#PCDATA)>
    <!ELEMENT SLOTS (RELATION+)>

    <!ELEMENT RELATION (ASSIGN_TO, FORM, CONTENT)>
    <!ATTLIST RELATION TYPE CDATA #REQUIRED>

    <!ELEMENT ASSIGN_TO (#PCDATA)>
    <!ELEMENT FORM (#PCDATA)>
    <!ELEMENT CONTENT (#PCDATA)>
\end{verbatim}
\end{example}
\normalsize

We use attributes to show the description of the concepts and we
can annotate the relevant relations between the concepts through
nested tags (e.g. the tag \emph{SLOTS}).

\begin{example} Excerpt from case frame analysis
\scriptsize
\begin{verbatim}
 <CONCEPT TYPE=Prozess>
    <WORD>Fertigen</WORD>
    <DESC>Schaffung von etwas</DESC>
    <SLOTS>
        <RELATION>
        <RESULT FORM="N(gen, fak) P(akk, fak, von)">
                fester Koerper</RESULT>
        <SOURCE FORM="P(dat, fak, aus)">aus formlosem
                Stoff </SOURCE>
        <INSTRUMENT FORM="P(akk, fak, durch)">durch
                Schaffen des Zusammenhalts</INSTRUMENT>
        </RELATION>
    </SLOTS>
</CONCEPT>
\end{verbatim}
\end{example}
\normalsize The example above is part of the result of the
analysis of the German phrase: {\em Fertigen fester Koerper aus
formlosem Stoff durch Schaffen des Zusammenhalts}\footnote{In
English: production of solid objects from formless matter by
creating cohesion}. The token {\em Fertigen} is classified as {\em
process} with the relations {\em source, result} and {\em
instrument}. The following phrases (noun phrases and preposition
phrases) are checked to make sure that they are assignable to the
relation requirements (semantic and syntactic) of the token {\em
Fertigen}.


\subsection*{Semantic interpretation of the syntactic
structure}

An other step to analyze the relations between tokens can be the
interpretation of the syntactic structure of a phrase or sentences
respectively. We exploit the syntactic structure of the
sublanguage to extract the relation between several tokens. For
example a typical phrase from an autopsy report: {\em Leber
dunkelrot.}\footnote{In English: Liver dark red.}

From semantic tagging we obtain the following information:
\begin{example} results of semantic tagging
\scriptsize
\begin{verbatim}
<CONCEPT TYPE="organ">Leber</CONCEPT>
<PROPERTY TYPE="color">dunkelrot</PROPERTY>
<XXX>.</XXX>
\end{verbatim}
\end{example}
\normalsize

In this example we can extract the relation "has-color" between
the tokens {\em Leber} and {\em dunkelrot}. This is an example of
a simple semantic relation. Other semantic relations can be
described through more complex variations. In these cases we must
consider linguistic structures like modifiers (e.g. \emph{etwas}),
negations (e.g. \emph{nicht}), coordinations (e.g.
\emph{Beckengeruest unversehrt und fest gefuegt}) and noun groups
(e.g. \emph{Bauchteil der grossen
\hyphenation{Koer-per-schlag-ader} Koerperschlagader}).

\section*{Current state and future work}

The XDOC document workbench is currently employed in a number of
applications. These include:

\begin{itemize}

\item knowledge acquisition from technical
documentation about casting technology

\item extraction of company profiles from WWW pages

\item analysis of autopsy protocols

\end{itemize}

The latter application is part of a joint project with the
institute for forensic medicine of our university. The medical
doctors there are interested in tools that help them to exploit
their huge collection of several thousand autopsy protocols for
their research interests. The confrontation with this corpus has
stimulated experiments with `bootstrapping techniques' for lexicon
and ontology creation.

The core idea is the following:

When you are confronted with a new corpus from a new domain, try
to find linguistic structures in the text that are easy to detect
automatically and that allow to classify unknown terms in a robust
manner both syntactically as well as on the knowledge level. Take
the results from a run of these simple but robust heuristics as an
initial version of a domain dependent lexicon and ontology.
Exploit these initial resources to extend the processing to more
complicated linguistic structures in order to detect and classify
more terms of interest automatically.

An example: In the sublanguage of autopsy protocols (in German) a
very telegrammatic style is dominant. Condensed and compact
structures like the following are very frequent:

\begin{quotation}
\noindent
\emph{Harnblase leer.}\newline \emph{Harnleiter frei.}
\newline \emph{Nierenoberflaeche glatt.}
\newline \emph{Vorsteherdruese altersentsprechend.} \newline \dots
\end{quotation}

These structures can be abstracted syntactically as
$<$Noun$>$$<$Adjective$>$$<$Fullstop$>$ and semantically as
reporting a finding in the form $<$Anatomic-entity$>$ has
$<$Attribute-value$>$ and they are easily detectable
\cite{roesner02}.

In our experiments we have exploited this characteristic of the
corpus extensively to automatically deduce an initial lexicon
(with nouns and adjectives) and ontology (with concepts for
anatomic regions or organs and their respective features and
values). The feature values were further exploited to cluster the
concept candidates into groups according to their feature values.
In this way container like entities with feature values like
`leer' (empty) or `gefuellt' (full) can be distinguished from e.g.
entities of surface type with feature values like `glatt'
(smooth).

\section*{Related Work}
The work in XDOC has been inspired by a number of precursory
projects:

In GATE \cite{GATESite,GATE} the idea of piping simple modules in
order to achieve complex functionality has been applied to NLP
with such a rigid architecture for the first time. The project LT
XML has been pioneering XML as a data format for linguistic
processing.

Both GATE and LT XML \cite{ltxml99} were employed for processing
English texts. SMES \cite{Neumann97} has been an attempt to
develop a toolbox for message extraction from German texts. A
disadvantage of SMES that is avoided in XDOC is the lack of a
uniform encoding formalism, in other words, users are confronted
with different encodings and formats in each module.

\section*{System availability}

Major components of XDOC are made publicly accessible for testing
and experiments under the URL:

 {\bf http://lima.cs.uni-magdeburg.de:8000/ }

\section*{Summary}

We have reported about the current state of the XDOC document
suite. This collection of tools for the flexible and robust
processing of documents in German is based on the use of XML as
unifying formalism for encoding input and output data as well as
process information. It is organized in modules with limited
responsibilities that can easily be combined into pipelines to
solve complex tasks. Strong emphasis is laid on a number of
techniques to deal with lexical and conceptual gaps and to
guarantee robust systems behaviour without the need for a priori
investment in resource creation by users. When end users are first
confronted with the system they typically are interested in quick
progress in their application but should not be forced to be
engaged e.g. in lexicon build up and grammar debugging, before
being able to start with experiments. This is not to say that
creation of specialized lexicons is unnecessary. There is a strong
correlation between prior investment in resources and improved
performance and higher quality of results. Our experience shows
that initial results in experiments are a good motivation for
subsequent efforts of users and investment in extended and
improved linguistic resources but that a priori costs may be
blocking the willingness of users to get really involved.

\bibliographystyle{acl}
\bibliography{coling}

\end{document}